\documentclass[letterpaper, 10 pt, conference]{ieeeconf}  

\IEEEoverridecommandlockouts                              

\usepackage{amsmath,amsfonts}
\usepackage{algorithm}
\usepackage{array}
\usepackage[caption=false,font=normalsize,labelfont=sf,textfont=sf]{subfig}
\usepackage{textcomp}
\usepackage{stfloats}
\usepackage{url}
\usepackage{verbatim}
\usepackage{graphicx}
\usepackage{cite}

\usepackage{threeparttable}

\usepackage{float}  

\usepackage{algpseudocode}

\title{\LARGE \bf
FFI-VTR: Lightweight and Robust Visual Teach and Repeat Navigation based on Feature Flow Indicator and Probabilistic Motion Planning
}

\author{Jikai Wang, Yunqi Cheng, and Zonghai Chen$^*$, ~\IEEEmembership{Senior Member,~IEEE,}
	\thanks{This work is supported by the National Natural Science
		Foundation of China (Grant No. 62103393). (Corresponding author: Zonghai Chen)}
	\thanks{Jikai Wang, Yunqi Cheng, and Zonghai Chen are with Department of Automation, University of Science and Technology of China (USTC), Hefei, 230027, PR China (e-mail: wangjk@ustc.edu.cn; chengyunqi@mail.ustc.edu.cn; chenzh@ustc.edu.cn).
	}%
	\thanks{$^*$Corresponding author. }
}

\begin{document}

\maketitle
\thispagestyle{empty}
\pagestyle{empty}

\begin{abstract}

Though visual and repeat navigation is a convenient solution for mobile robot self-navigation, achieving balance between efficiency and robustness in task environment still remains challenges. In this paper, we propose a novel visual and repeat robotic autonomous navigation method that requires no accurate localization and dense reconstruction modules, which makes our system featured by lightweight and robustness. Firstly, feature flow is introduced and we develop a qualitative mapping between feature flow and robot's motion, in which feature flow is defined as pixel location bias between matched features. Based on the mapping model, the map outputted by the teaching phase is represented as a keyframe graph, in which the feature flow on the edge encodes the relative motion between adjacent keyframes. Secondly, the visual repeating navigation is essentially modeled as a feature flow minimization problem between current observation and the map keyframe. To drive the robot to consistently reduce the feature flow between current frame and map keyframes without accurate localization, a probabilistic motion planning is developed based on our qualitative feature flow-motion mapping indicator. Extensive experiments using our mobile platform demonstrates that our proposed method is lightweight, robust, and superior to baselines. The source code has been made public at https://github.com/wangjks/FFI-VTR to benefit the community.

\end{abstract}

\section{INTRODUCTION}

Recent years have witnessed the rapid development of mobile robots, such as robotic dogs and humanoid robot. To be deployed in multiple task environments, mobile robot self-navigation is highly required. With the help of artificial intelligent methods \cite{vaswani2017attention}, the recent proposed mobile robot navigation methods have demonstrated significant intelligence \cite{zhang2025mapnav,chen2022think}, powerful trajectory planning and control ability \cite{ren2025safety,le2024comprehensive}, and global exploration ability \cite{raj2024intelligent,xu2024pare}. However, the computational cost and the limited environmental adaptability of these navigation systems make it challenged to deploy mobile robots in multiple tasks environments efficiently. As a fundamental component, navigation should be lightweight and thus more computational resource can be used for high-level tasks. Furthermore, most navigation systems are constructed based on accurate localization system, which has corner cases in multiple environments.


Visual Teach and repeat navigation (VTR) \cite{van2024visual,10611662} has been proved to be an effective solution for fast mobile robot deployment. In teaching phase, human operator guides the robot following the task path using remote control. At the same time, the navigation system records the visual observation and motion information as a map. In  repeating phase,  robot compares its current observation with the map and then decides its next motion to repeat the task path as close and complete as possible. To achieve VTR navigation, first kind of methods are constructed based on accurate localization module \cite{paul2024mpvo}. With known global metric pose estimation, VTR navigation is an easy problem. However, highly stable and adaptable localization module remains challenged and an open problem \cite{9440682}.  The second kind of methods \cite{furgale2010visual,dall2021fast, 10610234} are constructed based on relative or topological localization. They compute relative pose within the local map and formulate an optimized motion planning to increase its consistency with the teaching map. Thus, in the existing methods, the cores are to retrieve visual cues accurately and robustly, and correct the motion based on the retrieval results. The visual retrieving module should be robust to environment dynamic changing, illumination changing, and large viewpoint changes. However, though robustness to environment appearance changes has been solved to some extent, most existing methods are based on an assumption that the robot are always on the teaching path and the dynamic objects are neglected.

In this paper, we aim to construct a lightweight and robust VTR navigation system, which enables the robot to be deployed in dynamic and various environments. Our system is based on one knowledge that the pixel flow can be used for indicating camera motion. Since dense pixel flow computation is sensitive to viewpoint, thus, we extract deep learning-based features, in which we use xfeats \cite{potje2024xfeat} feature, and compute feature flow. In the  teaching process, the map is constructed as a lightweight keyframe graph and feature flows between adjacent keyframes are preserved, in which feature flow is defined as feature pixel location distance between matched features. In    repeating process, the system keeps tracking the map keyframe and compute feature flows. A probabilistic motion planner considering both feature flow indicator and obstacles are proposed to drive the robot.  Our contributions are as follows. 

\begin{itemize}
	\item A lightweight and robust visual teach and repeat navigation system is proposed, which can be deployed in complex and dynamic environments. No metric localization module is required.
	\item Feature flow-based motion indicator is proposed. The navigation is totally driven by the feature flow, which makes our system robust and lightweight.
	\item Feature flow-based probabilistic motion planning module is proposed, which demonstrates that navigation can be achieved without based on fully sufficient information. We have made our source code open.
\end{itemize}

\section{Related Work}

  Inspired by human that the environment is expressed as a combination of actions and observations, a concept of Scene Action Maps (SAMs) is proposed in SAM \cite{loo2024scene},  which represents the environment as a graph of interconnected navigational behaviors. The SAMs enable robots to navigate effectively even with limited metric and spatial information.  In\cite{cuizhu2023one}, the teaching image sequence is divided according to navigation actions. Instead of representing the environments as a topological graph of visual frames, Garg et al. \cite{garg2024robohop}  construct graph of segments and establish a novel mechanism for intra- and inter-image connectivity based on segment-level descriptors and pixel centroids. Their method are more friendly to object navigation. In \cite{10433735}, critical navigation waypoints are extracted and the mapping between waypoints visual cues and navigation cues are learned. \cite{furgale2010visual} is the most classical VTR framework, which constructs overlapping submaps motivated by drift errors of visual odometry. Then, the robots continuously perform submap localization instead of global localization. Path tracking is achieved based on a ground plane assumption.  \cite{paul2024mpvo} also argues that Visual Odometry (VO) is not globally accurate thus cannot be directly used for point goal navigation. To alleviate this problem, they fuse a traditional visual odometry for coarse pose estimation and a learning-based visual odometry for fine pose correction.   \cite{van2024visual} is  constructed based on the viewpoint that visual odometry is with drift errors and not be able to support global navigation. Thus, they combine VO and periodic orientation correction using visual comparison, which ensure that the agent can follow back its route.
\cite{10578334} introduces a predictive approach to data acquisition, where the robot uses its past experiences and current observations to make informed decisions about which data to collect.
\cite{rozsypalek2023multidimensional} considers that the changing environment makes it ambiguity  to recognize places. They use particle filter to simultaneously estimate the frame correspondence of current observation on teaching map.  In \cite{krajnik2018navigation}, a global localization-free simple visual teach-and-repeat navigation system is constructed, in which the robot only correct its heading based on frame matching results during the repeating phase.

\section{Proposed Method}

\subsection{System Overview}
\begin{figure}[t]
	\centerline{\includegraphics[width=0.85\columnwidth]{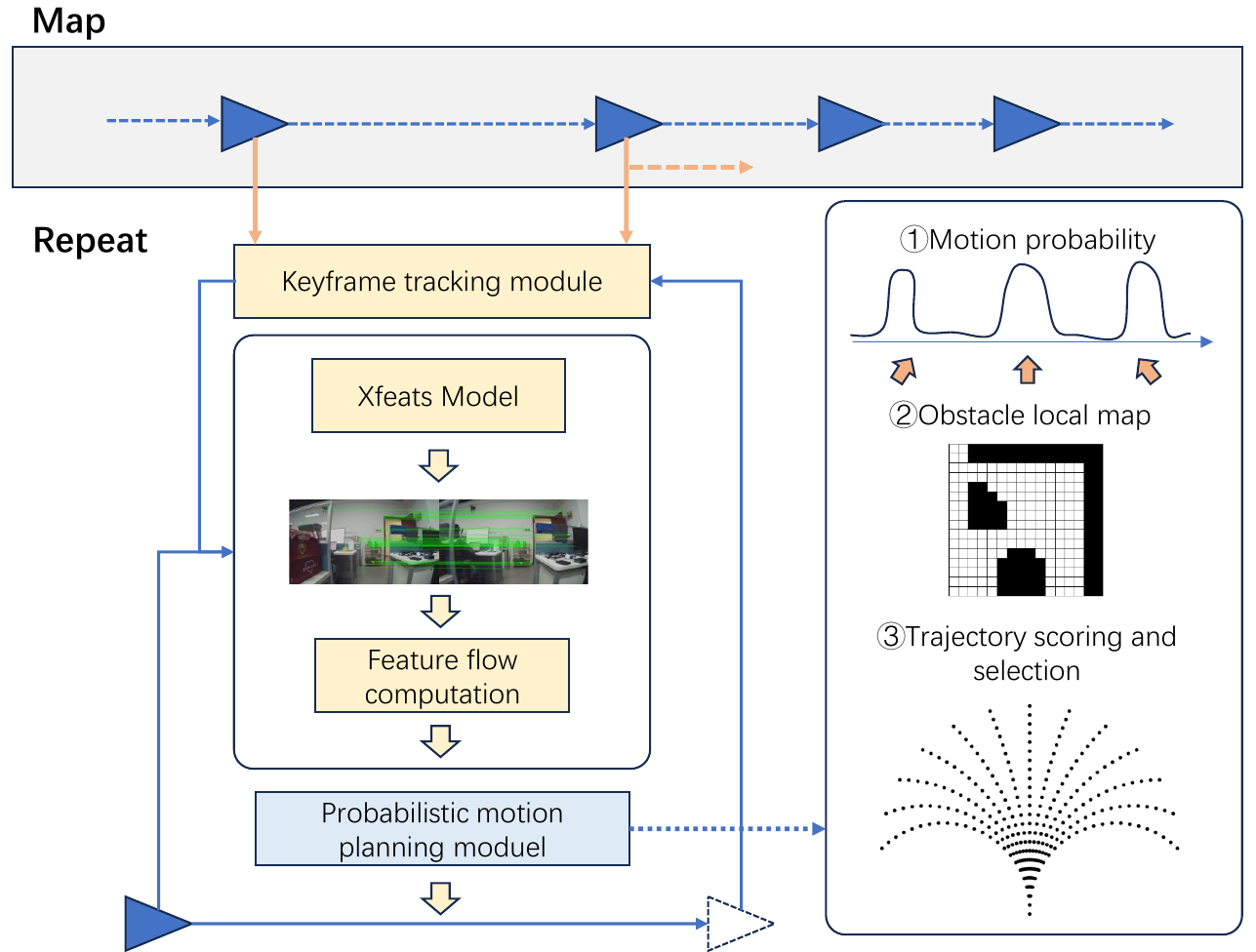}}
	\caption{Our system flowchart.}
	\label{systemframework}
	\vspace{-0.5cm}
\end{figure}

Our system is constructed based on the findings that feature flow can be motion indicator. And VTR navigation can be essentially regarded as a task to minimization the feature flow between repeating path and teaching path. It is consists of mapping and navigation parts. In the mapping process, the robot sequentially collects visual images and decide keyframes according to feature flow. The map is formulated as a keyframe graph, in which the edges preserve feature flow between adjacent keyframes. In the navigation process, the system firstly performs place recognition and compute feature flow with the matched keyframe. Then, a feature flow window involving the next keyframe is constructed and a probabilistic motion planning module is activated to output optimal movement, in which obstacle avoidance function is also incorporated. System framework illustration is presented in Fig. \ref{systemframework}.

\subsection{Feature flow-based motion indicator}

Computing 6-Dof pose according to visual images has been extensively studied in SLAM community. However, most visual slam systems are sophisticated with high computing cost. Actually, in navigation process, we notice that highly accurate pose estimator is not really required. Optical flow is a simple cue that is highly related to camera motion. Thanks to the recent proposed deep learning-based visual features with high discrimination ability, we propose feature flow that is more robust to viewpoint change compared with optical flow. Specifically, let $\mathcal{I}_q$ and $\mathcal{I}_r$ denote the query image and reference. In the local coordinates of camera pose of $\mathcal{I}_r$, the camera pose of $\mathcal{I}_q$ is denoted as $[\mathbf{R},\mathbf{t}]$. Then, let $\mathbf{p} = [x,y,z]^T$ denote a feature's 3D position and its projection on  $\mathcal{I}_q$ and $\mathcal{I}_r$ are $\mathbf{u}^q$ and $\mathbf{u}^r$, respectively. Then, according to visual re-projection model
\begin{equation}
	\mathbf{u}^r = \begin{bmatrix}
		u\\v\\1
	\end{bmatrix} = \frac{1}{z}\mathbf{K}\mathbf{p} = \begin{bmatrix}
	 f_x \frac{x}{z} + c_x\\
	 f_y \frac{y}{z} + c_y\\
	 1
	\end{bmatrix},
\end{equation}
where $\mathbf{K}$ is the intrinsic matrix. Given the mapping function, compared with $\mathcal{I}_r$, if the camera moves straight $\delta$ m, it means that $\mathbf{R}$ is identical matrix and $\mathbf{p}$ is turned to $ [x,y,z-\delta]^T$ Then, we have
\begin{equation}
	\mathbf{u}^q
	= \begin{bmatrix}
		f_x & 0 & c_x \\
		0 & f_y & c_y \\
		0 & 0 & 1
	\end{bmatrix}
	\begin{bmatrix}
		\frac{x}{z - \delta} \\
		\frac{x}{z - \delta} \\
		1
	\end{bmatrix}
	= \begin{bmatrix}
		f_x \frac{x}{z - \delta} + c_x \\
		f_y \frac{y}{z - \delta} + c_y \\
		1
	\end{bmatrix}.
\end{equation}
Then, the pixel location variation is
\begin{equation}
	\Delta\mathbf{u} = \mathbf{u}^q-\mathbf{u}^r = \begin{bmatrix}
		 f_x\cdot x \cdot\frac{\delta}{z (z - \delta)}\\
		 f_y\cdot y \cdot \frac{\delta}{z (z - \delta)}\\
		 0
	\end{bmatrix}.
\end{equation}
Since the robot is navigating on the ground plane, we mainly focus on the first dimension of the pixel location variation, which is $f_x\cdot x \cdot\frac{\delta}{z (z - \delta)}$. Its positive or negative sign is totally determined by $x$. Considering all the pixels distributed on the image plane, we have
\begin{equation}
	\sum_i\Delta\mathbf{u}_i[1] = \sum_i f_x\cdot x_i \cdot\frac{\delta}{z_i (z_i - \delta)} \to 0,
\end{equation}
due to that these pixels' $x$ values are evenly distributed on the negative and positive $x$-axis. Thus, we have a following qualitative rule:

\textbf{Qualitative rule 1}: If robot moves forward straight, then the sum of pixel location's signed variation along $x$-axis is around 0. If the robot is static, the sum is 0.

Notice that it is a qualitative rule due to that we neglect the difference of pixels' $z$ values. 

If the robot spins around the $y$-axis of camera coordinates 
and $\mathbf{t}$ is zero vector. Then, we have
\begin{equation}
\mathbf{u}^q
= \frac{1}{z}\mathbf{K} \mathbf{R}\mathbf{p}
= \begin{bmatrix}
	f_x \frac{x \cos\theta + z \sin\theta}{z} + c_x \\
	f_y \frac{y}{z} + c_y \\
	1
\end{bmatrix}.
\end{equation}
Notice that we assume that the depth change is neglected.
Then, the pixel location variation caused by rotational movement is
\begin{equation}
	\Delta\mathbf{u} = \mathbf{u}^q-\mathbf{u}^r \approx \begin{bmatrix}
		f_x\sin(\theta),
		0,
		0
	\end{bmatrix}^T.
\end{equation}

It means that the pixels' location variations have the same positive or negative sign determined by $\theta$. Furthermore, their absolute values are proportional to $\theta$. Then, we have the second rule:

\textbf{Qualitative rule 2}: If the robot rotates to left, then the caused pixels' location variation is a positive value. If the  robot rotates to right, then the caused pixels' location variation is a negative value. 

Thus, according to the above rules, the distance between $\mathbf{u}^q$ and $\mathbf{u}^r$ can be a qualitative indicator of the relative transformation $\mathbf{R}$ and $\mathbf{t}$. Complex movements can be regarded as the combination of straight-line motion and rotational motion.

Since it is hard to match all the pixels, especially in large viewpoint change, we use xfeats model to extract features on the two images and perform feature matching according to feature descriptor distances.  The matched feature set is denoted as $\mathcal{F}_{q|r}=\{(\mathbf{u}_i^q,\mathbf{u}_i^r),i=1,\cdots,N\}$. Then, feature flow is computed as
\begin{equation}
	\mathbf{f}_{q|r} = \frac{1}{N}\sum_{i=1}^{N}(\mathbf{u}_i^r-\mathbf{u}_i^q).
\end{equation}

\begin{figure}[t]
	\centerline{\includegraphics[width=0.9\columnwidth]{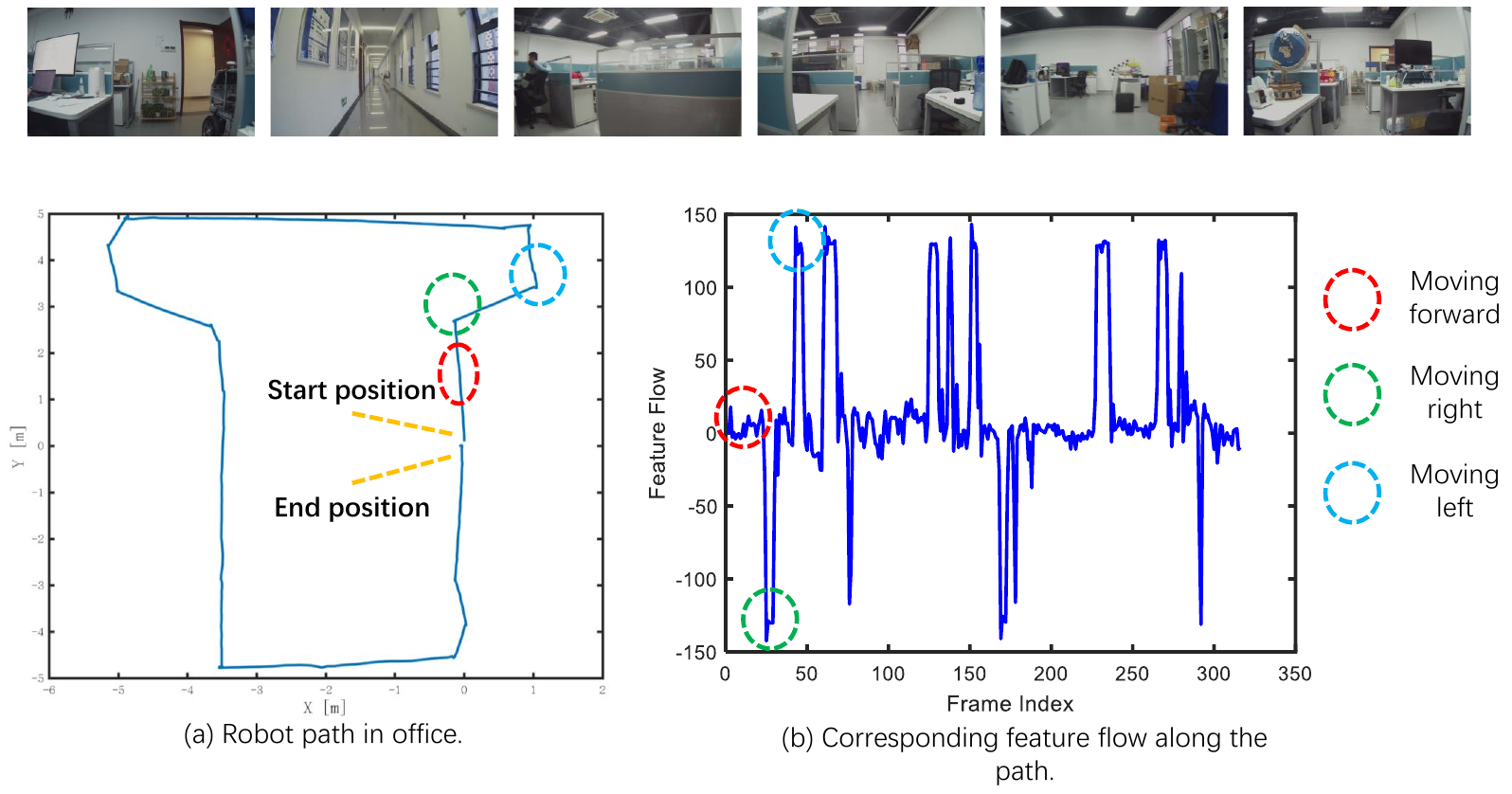}}
	\caption{Feature flow-based motion indicator illustration. (a) is the robot moving path in office environment. (b) presents the feature flow of the keyframes. We can tell that the variance of feature flow is highly related with the robots' motion. Feature flow is around 0 when robot is moving forward.  Feature flow is a large positive value when robot is moving rotational to left.  Feature flow is a large negative value when robot is moving rotational to right.}
	\label{featureflow}
	\vspace{-0.5cm}
\end{figure}

Since our method is designed for robots navigating on ground plane, the feature flow is along the x-axis of image plane. Thus, in this paper, we use $f_{q|r} = \mathbf{f}_{q|r}[0]$ represent feature flow. If $\mathcal{I}_q$ and $\mathcal{I}_r$ are collected from the same pose, then $f_{q|r}$ equals to 0. If  $f_{q|r}$ is around 0, it means that the robot motion between $\mathcal{I}_q$ and $\mathcal{I}_r$ is most likely going straight. If $f_{q|r}$ is a large positive number,  it means that the robot motion is most likely going to left. If $f_{q|r}$ is a large negative number,  it means that the robot motion is most likely going to right. Since the matched features are not evenly distributed on the image plane, thus the feature flow-based motion indicator is probabilistic. An experimental illustration is presented in Fig.  \ref{featureflow}.

\subsection{Mapping in Teaching}

The aim of visual teaching is to formulate the environmental map, during which the task path is under the guidance of human operator. To construct a lightweight map, we extract keyframe from the visual image stream. Specifically, the first frame is set as initial keyframe. When a new frame comes, we extract xfeats features and perform feature matching with the last keyframe. Feature flow is then computed and a new keyframe is created if feature flow is beyond a threshold. The map is finally represented as
\begin{equation}
	\mathcal{M} = \{\mathcal{K}_i = [\mathcal{I}_i,\mathcal{F}_i,f_{{i+1}|i}],i=1,\cdots,K\},
\end{equation}
where $\mathcal{F}_i$ denote the extracted xfeats features, including feature points and descriptors. $f_{{i+1}|i}$ denotes the feature flow between the $i$-th keyframe and $(i+1)$-th keyframe. Our map is lightweight and easy to be extended. In $\mathcal{M}$, the relative motion between adjacent keyframe is encoded in $f_{{i+1}|i}$. Once $\mathcal{I}_i$ is retrieved, we can infer the movement to $\mathcal{I}_{i+1}$ according to $f_{{i+1}|i}$. 

\subsection{Navigation in Repeating}
Repeating navigation is to follow the mapping path. The initial positions are the same. Then, we perform map keyframe tracking using Kalman Filter framework, and computing the optimal instant movement decision based on probability defined by feature flow. The details are presented as follows. 

\subsubsection{Place recognition}

Place recognition is essentially to keep tracking of the keyframe in map. Though DBoW2 is used for place recognition and widely applied in VTR navigation systems, it is sensitive to viewpoint changes. In this paper, we propose an iteratively keyframe tracking mechanism. Let $\mathcal{I}_t$ denote the current observing image and $\mathcal{K}_l$ is the currently tracked keyframe. For new image $mathcal{I}_{t+1}$, we perform feature matching with $\mathcal{K}_l$ and $\mathcal{K}_{l+1}$ sequentially. Then, the keyframe whose has the largest matched feature's number is updated as the new tracked keyframe. During the tracking process, to avoid lost tracking, we set a threshold for the number of matched features between current image with tracked keyframe. If it is below the threshold, then a locally loop detection is activated. Instead of detecting loops on the whole map, we only perform loop detection on the local keyframe set.

\subsubsection{Movement probability computing based on feature flow window}

Given the proposed qualitative feature flow-based motion indicator, we compute the instant movement probability according to the place recognition results. Specifically, for the current $i$-th frame, its looped keyframe is $\mathcal{K}_l$, then we compute $f_{l|i}$ and $f_{l+1|i}$.

 We have three movement events, which are moving forward straight $\{0\}$, moving to left $\{1\}$, and moving to right $\{2\}$. Then, we compute a fake probability of each event $E_j$ as follows
\begin{equation}
	\label{prob}
	\begin{aligned}
		p(E = 0) &= \sum_{n=0}^{1} p(E=0|f_{l+n|i}) 
		=  \sum_{n=0}^{1}\omega_n \exp(-\frac{f_{l+n|i}^2}{2\sigma^2}),\\
		p(E = 1) &= \sum_{n=0}^{1}\omega_n \mathbb{I}(f_{l+n|i}>0) \left(1-\exp(-\frac{f_{l+n|i}^2}{2\sigma^2})\right) ,\\
		p(E = 2) &= \sum_{n=0}^{1}\omega_n\mathbb{I}(f_{l+n|i}<0) \left(1-\exp(-\frac{f_{l+n|i}^2}{2\sigma^2})\right),\\
	\end{aligned}
\end{equation}
where $\mathbb{I}(\cdot)$ returns to 1 if its condition is satisfied, otherwise, 0 is returned. 
\begin{equation}
	\label{weight}
	\omega_n = \exp(-\frac{n^2}{2\sigma_w^2}).
\end{equation}
The application of  $\omega_n$ enables that $p(E_j)$ pays more attention on the cloest keyframe's information. $\mu_j$ is the expectation of the feature flow corresponding to $E_j$. $\sigma$ and $\sigma_w$ are preset standard variances. Under the condition of each feature flow, we sum their fake probability. By normalizing the computed probability values, probabilistic distribution of the three movements is derived. Considering consecutive keyframe's movements can effectively help the navigation system get out of local optimal. 

\subsubsection{Obstacle avoidance fusion}

In our navigation system, the robot is driven by feature flow and obstacle avoidance. We construct an effective method based on candidate trajectory selection to fuse the two factors.

In order to provide decision basis for motion planning and real-time control, we generate a series of trajectory candidate sets satisfying motion constraints by sampling based on the robot kinematics model offline. The sampling process takes into account the robot's dynamic characteristics, turning radius, and maximum speed and acceleration limits to ensure the feasibility of trajectory generation.
 An illustration of the generated trajectory candidate set is presented in Fig.  \ref{platform}. Notice that the trajectories are in the body coordinates of the robot and each candidate corresponds to pre-defined velocity and angular command.

Considering dynamic and static obstacles, we also make full use of the two-dimensional single-line LiDAR mounted on the robot platform to obtain real-time environmental information for obstacle avoidance. We construct a local grid map and labeling each grid free or occupied according to the LiDAR range measurements.  After the occupied map is constructed, the occupied grids are flatted according to the robot body size and safety margin.  Then, the grids occupied by the trajectory candidates and the obstacles are cross-checked to determine whether each trajectory candidate collides with the occupied grid after expansion. Any trajectory candidate that intersects with obstacles are immediately eliminated, ensuring the safety of path planning.

After filtering trajectory candidates according to grid occupancy, we perform trajectory candidates scoring according to the movement probability derived according to feature flow. Specifically, if the maximum probability of event is moving straight, then the local goal in the robot's coordinates is set as $[1,0]$. For moving left, the goal is set as $[1,1]$. For moving right, the goal is set as $[1,-1]$. Our motion planning module is probabilistic due to that the provided goal point is qualitative and probabilistic.

In the coordinate system of the robot, the angle between goal point and $x$-axis is defined as $\theta_g$, and the angle between the end of the path and $x$-axis is defined as $\theta_p$. The scoring formula for each candidate $P_i$ is 
\begin{equation}
	S(P_i)=1-\sqrt{\sqrt{0.005\cdot\mathrm{\theta_a}}},
\end{equation}
where $\theta_a=\left | \theta_g- \theta_p \right |$.
Each path in the set of feasible paths is scored, and the path with the highest score is the optimal path. According to the trajectory group of the optimal path, the steering angular speed is determined and published to drive the robot automatically.

\section{Experiments}

\subsection{Mobile Robot Configuration}
\begin{figure}[t]
	\centerline{\includegraphics[width=0.95\columnwidth]{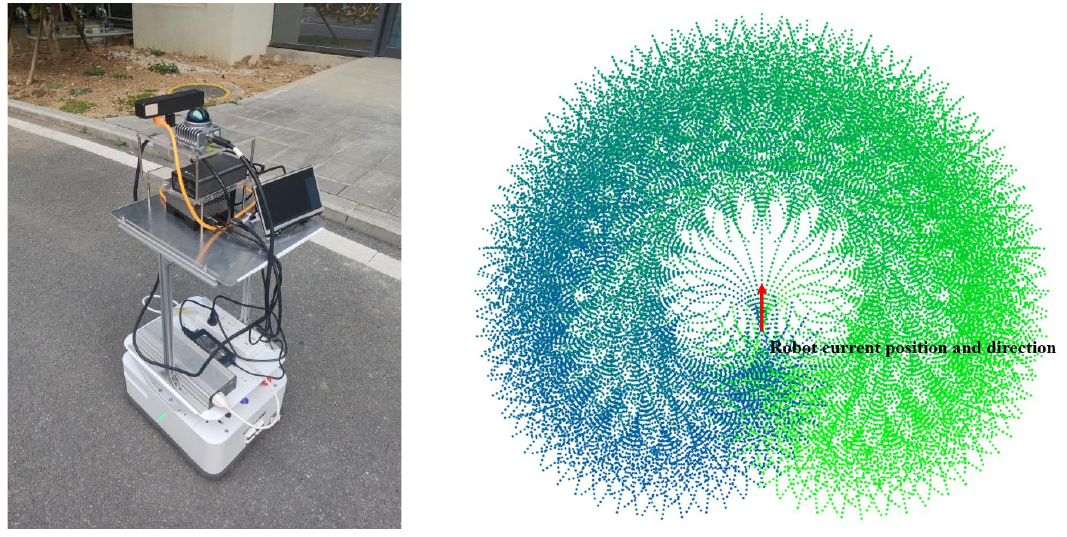}}
	\caption{Our mobile platform. An IMU-embedded stereo camera and 3D LiDAR sensor are mounted for environmental perception and localization. }
	\label{platform}
	\vspace{-0.3cm}
\end{figure}

Our robot is constructed based on a differential driving platform. A IMU-embedded camera (MYNTEYE-SC) and LiDAR-IMU sersor (Livox Mid-360) are mounted. We implement iG-LIO \cite{10380742} using the LiDAR-IMU sersor observations to record the robot's trajectories for evaluation only. Onboard computer is NVIDIA Jetson AGX Orin. The indoor mobile platform is presented in Fig. \ref{platform}.

\subsection{Experimental Configuration}
\begin{figure}[t]
	\centerline{\includegraphics[width=0.95\columnwidth]{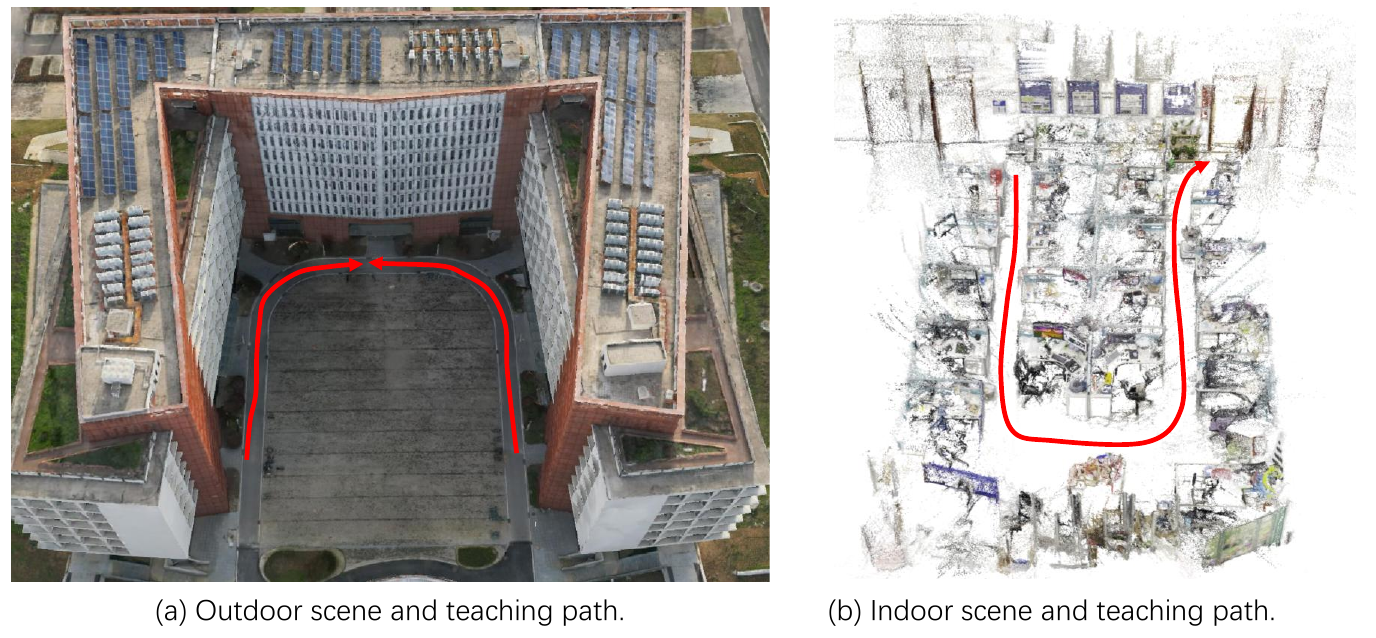}}
	\caption{Our experimental environments. We have two teaching path segments in outdoor scene and one teaching path segments in indoor scene, which are denoted in red lines.}
	\label{map}
	\vspace{-0.5cm}
\end{figure}

We implement experiments in indoor and outdoor dynamic scenarios, which are shown in Fig. \ref{map}. Indoor scenario is our lab office and the walkway is narrow. In outdoor scenario, there exists dynamic objects, illumination changes, and viewpoint changes. $\sigma$ in Eq. (\ref{prob}) and Eq. (\ref{weight}) are set to 20 and 2, respectively. We firstly collect images to construct map along the teaching trajectory. Then, we implement visual repeating navigation.

\subsection{Baselines and Evaluation}

 We compare our method with the most representative VTR methods referring to \cite{dall2021fast}, termed as QVTR\footnote{https://github.com/QVPR/teach-repeat}, and our reproduced verison of VT\&R3\footnote{https://github.com/utiasASRL/vtr3} \cite{furgale2010visual}. Notice that both baselines require accurate visual odometry module. Since these methods do not have obstacle avoidance module, we incorporate our local navigation module into their systems. 
 
 In this paper, the VTR is defined as success if the robot can navigate to the end point from the start point, no matter how much deviation from the teaching route during the repeating process. Thus, in this paper, the distance between the real arrived end point and the preset end point of teaching route is used as evaluation metric, termed as end-point distance.
 
\subsection{Experimental Results}

\begin{figure}[t]
	\centerline{\includegraphics[width=0.95\columnwidth]{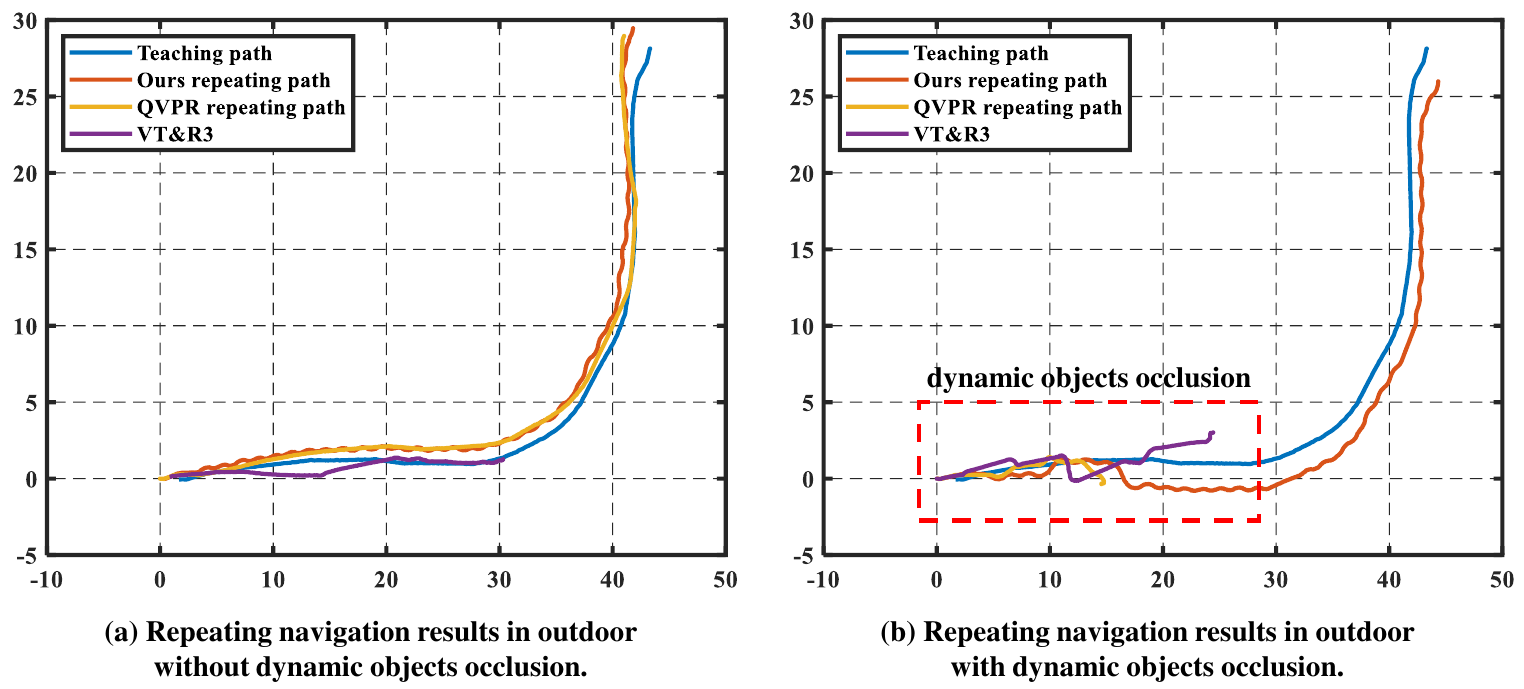}}
	\caption{Repeating navigation results of the baselines on the first teaching segment in outdoor scene. }
	\label{1st_com}
	\vspace{-0.2cm}
\end{figure}

\begin{figure}[t]
	\centerline{\includegraphics[width=0.95\columnwidth]{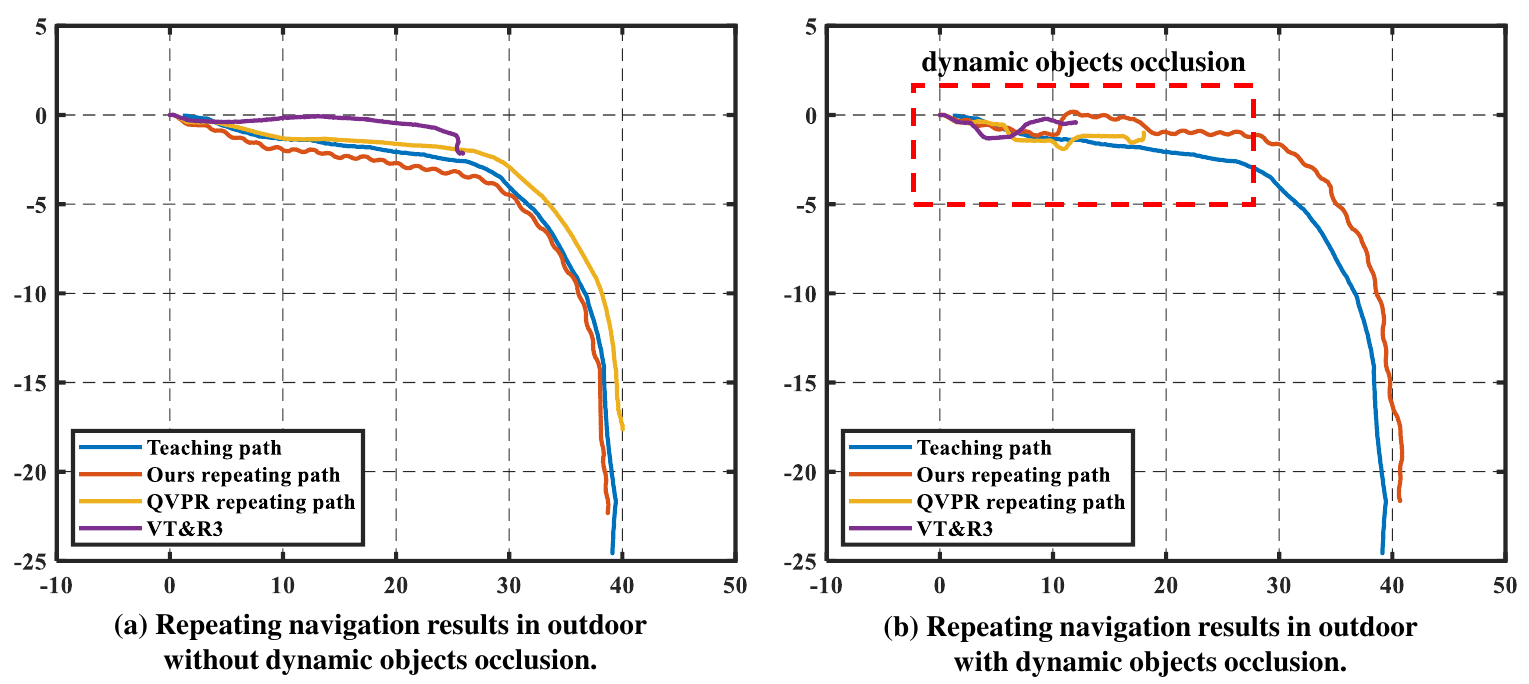}}
	\caption{Repeating navigation results of the baselines on the second teaching segment in outdoor scene. Robot start at $(0,0)$ position.}
	\label{2st_com}
	\vspace{-0.2cm}
\end{figure}

\begin{figure}[t]
	\centerline{\includegraphics[width=0.5\columnwidth]{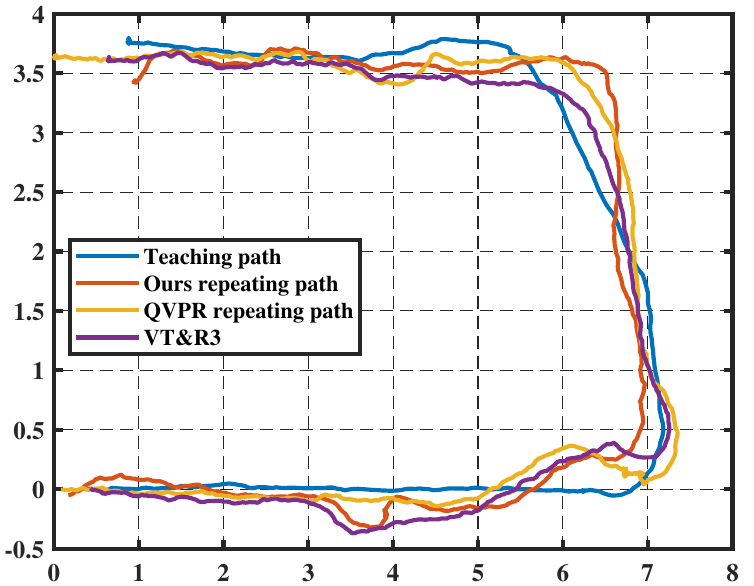}}
	\caption{Repeating navigation results of the baselines in indoor office scene. }
	\label{officecom}
	\vspace{-0.2cm}
\end{figure}

\begin{figure}[t]
	\centerline{\includegraphics[width=0.85\columnwidth]{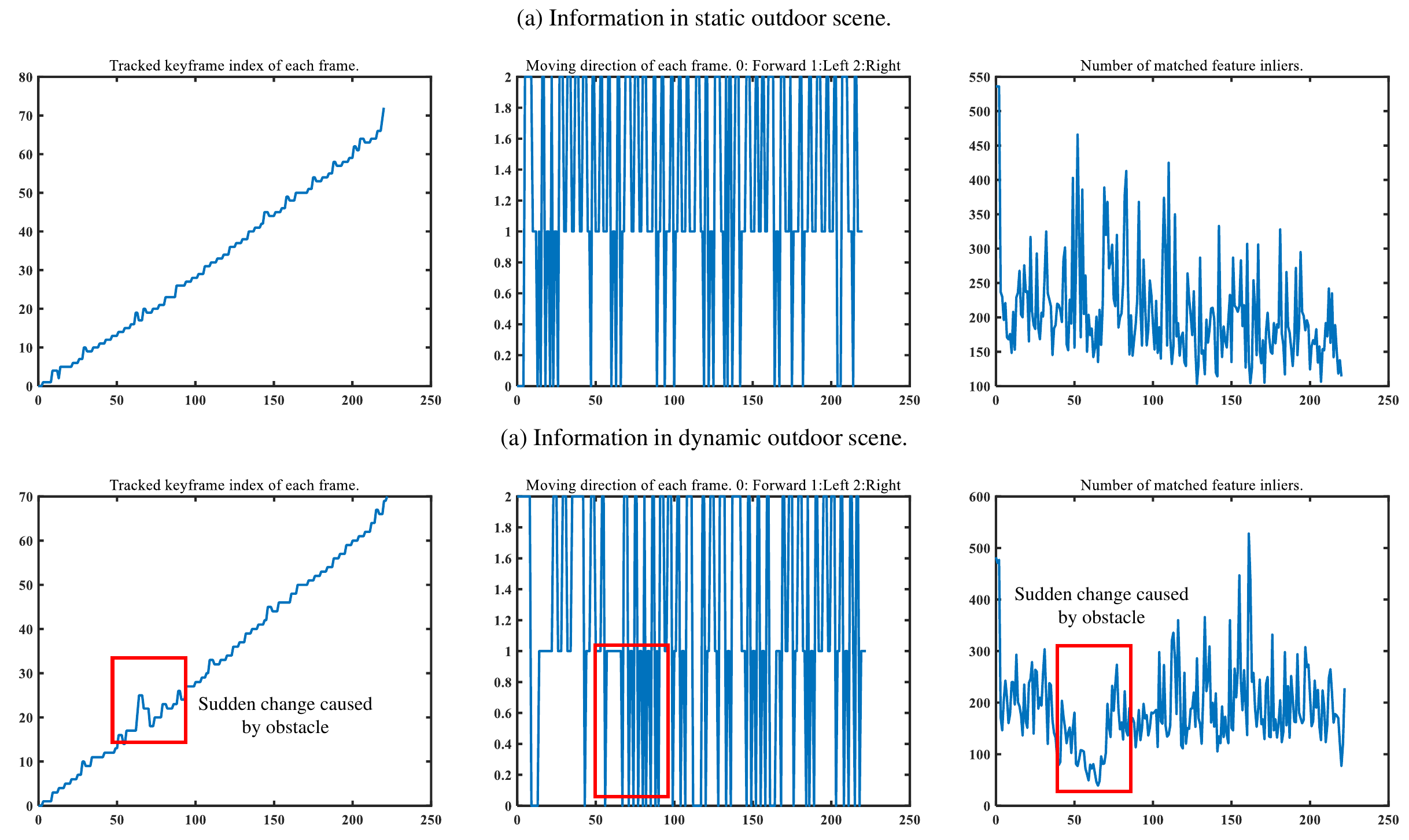}}
	\caption{The tracked frame index, moving direction, and feature inlier number of each frame during repeating navigation process are presented. In the second row, we can clearly tell the affects of dynamic objects occlusion. }
	\label{keyframetracked}
	\vspace{-0.5cm}
\end{figure}

\begin{figure}[t]
	\centerline{\includegraphics[width=0.85\columnwidth]{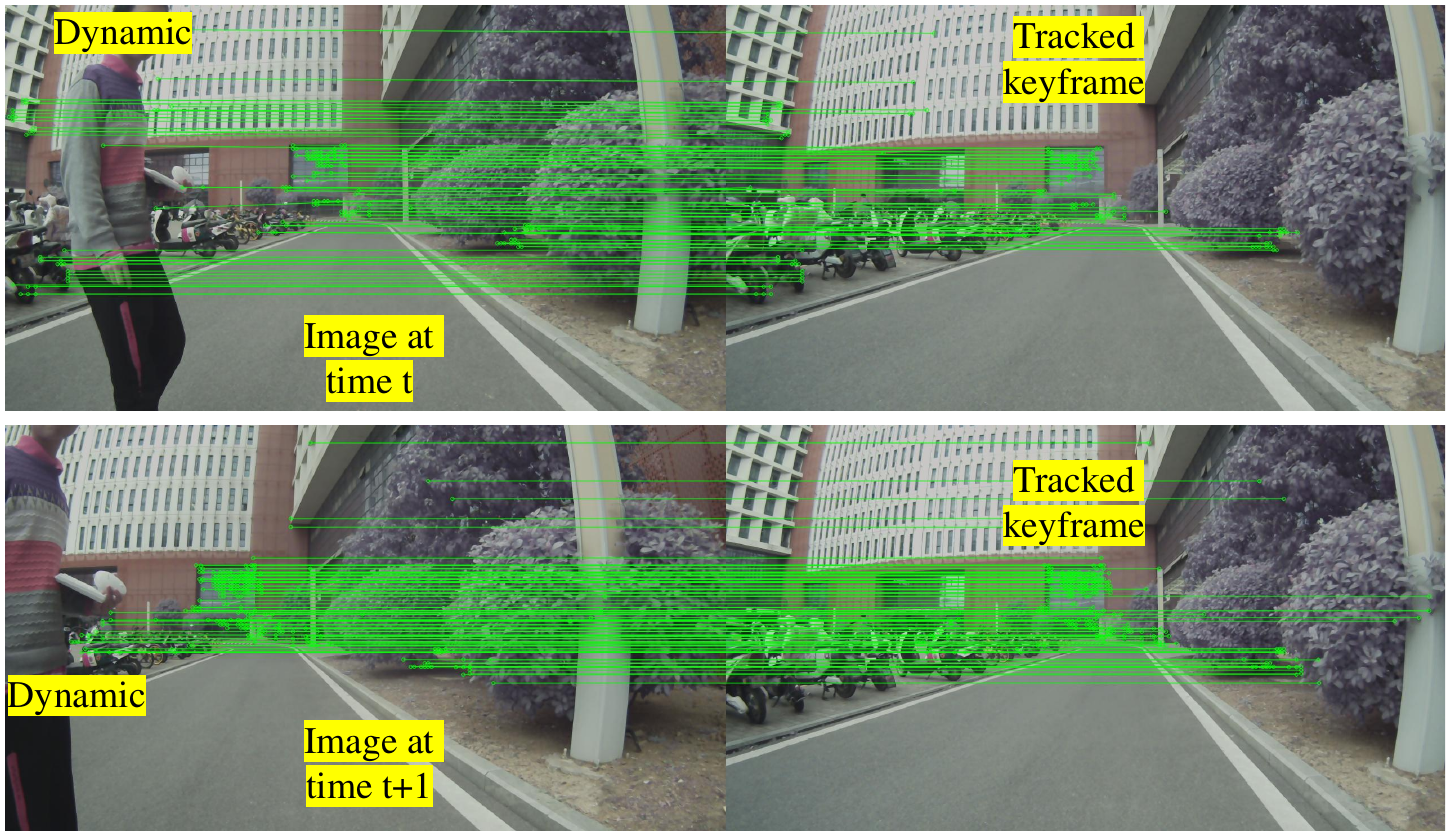}}
	\caption{Presentation of the dynamic object avoidance process of our system. }
	\label{dynamic}
	\vspace{-0.2cm}
\end{figure}

The repeating navigation results in outdoor scene are shown in Fig. \ref{1st_com} and \ref{2st_com}. When the scene is static, ours and QVPR can complete the repeating navigation, while VT\&R3 fails around the corners due to goal points computed wrongly. When dynamic people exist and block the path, only our method can successfully avoid obstacle and continue repeating navigation. Both baseline methods lose tracking of the map. Dominate reason is that our method is based on robust sparse feature matching. QVPR relies on global image matching, which is sensitive to viewpoint changes. VT\&R3 heavily relies on 3D-2D feature reprojection for place recognition, thus is also sensitive to viewpoint changes due to object occlusion. It should be also noticed that out repeating trajectory frequently adjusts its local directions. In the testing environment, the extracted feature is sparse while our feature flow indicator is based on the assumption of features are abundant, thus the computed feature flow is biased compared with the moving motion. Our probabilistic motion planning module is highly robust to such condition. The end point distances are presented in Table. \ref{endpoint}. The repeating results in office scene is presented in Fig. \ref{officecom}. All methods demonstrate success navigation. In general, our method is superior to the baselines. The middle results of our method in outdoor scene with and without dynamic objects are presented in Fig. \ref{keyframetracked}. Around the 60-th frame, we can tell a sudden change of the matched keyframe and the corresponding feature inliers' number is suddenly decreased, which is caused by the dynamic object occlusion. After obstacle avoidance finishes, these information return to normal state, which demonstrates that our keyframe tracking mechanism is effective.  Dynamic object avoidance process of our method is presented in Fig. \ref{dynamic}. We can see the turning right action of robot due to obstacle avoidance. Our system can run in real-time when the robot moves in around 2 m/s, and the most computational cost step is to compute the xfeats feature matching.

\begin{table}[t]
	\centering
	\label{endpoint}
	\caption{End point distance of the baselines.}
	\begin{tabular}{|c|cccc|c|}
		\hline
		& \multicolumn{4}{c|}{Outdoor}                                                                                                                                                                                                                                                                      & Indoor                      \\ \hline
		& \multicolumn{1}{c|}{\begin{tabular}[c]{@{}c@{}}path1\\ static\end{tabular}} & \multicolumn{1}{c|}{\begin{tabular}[c]{@{}c@{}}path1\\ dynamic\end{tabular}} & \multicolumn{1}{c|}{\begin{tabular}[c]{@{}c@{}}path2\\ static\end{tabular}} & \begin{tabular}[c]{@{}c@{}}path2\\ static\end{tabular} & \multicolumn{1}{l|}{static} \\ \hline
		QVTR  & \multicolumn{1}{c|}{2.42}                                                   & \multicolumn{1}{c|}{40.46}                                                   & \multicolumn{1}{c|}{7.01}                                                   & 31.61                                                  & 0.88                          \\ \hline
		VT\&R3 & \multicolumn{1}{c|}{29.86}                                                  & \multicolumn{1}{c|}{31.42}                                                   & \multicolumn{1}{c|}{25.99}                                                  & 36.29                                                  & \textbf{0.28}                          \\ \hline
		Ours  & \multicolumn{1}{c|}{\textbf{1.99}}                                          & \multicolumn{1}{c|}{\textbf{2.38}}                                           & \multicolumn{1}{c|}{\textbf{2.28}}                                          & \textbf{3.32}                                          & 0.36                          \\ \hline
	\end{tabular}
	\vspace{-0.5cm}
\end{table}

\section{CONCLUSIONS}
In this paper, we develop a novel lightweight visual teach and repeat system based on feature flow motion indicator. No metric localization is required and our system can avoid obstacles using probabilistic mechanism. We prove that visual navigation can be achieved using such framework. Though we apply xfeats for robust feature matching, keyframe tracking lost still exists in extremely challenging environment. Furthermore, matched xfeat features are sparse and cannot be ensured to distribute on image plane evenly, which may lead to inappropriate  motion indication. Thus, in the future, we will introduce instance segmentation for robustly flow-motion computing and embodied intelligent models for robustly decision, especially for recovering navigation after getting lost.

\end{document}